\documentclass[preprint]{imsart}

\usepackage{amsbsy}
\usepackage{amsmath}
\usepackage{amsthm}

\usepackage{color}
\usepackage[all]{xy}

\usepackage{wrapfig}

\usepackage{epsf}
\usepackage{epsfig}

\usepackage{bbm}

\usepackage[vlined,linesnumberedhidden,ruled,resetcount]{algorithm2e}

\SetKwBlock{Repeat}{repeat}{}
\SetKwInOut{Initialization}{Initialization}

\newcommand{\bfX}{\boldsymbol{X}}
\newcommand{\bfW}{\boldsymbol{W}}

\newcommand{\bfC}{\boldsymbol{C}}
\newcommand{\ci}{\perp\!\!\!\perp}
\newcommand{\nci}{\not\perp\!\!\!\perp}

\newcommand{\bfBeta}{\boldsymbol{\beta}}

\def\ind{\mathbbm{1}}

\newcommand{\beginsupplement}{%
        \setcounter{table}{0}
        \renewcommand{\thetable}{S\arabic{table}}%
        \setcounter{figure}{0}
        \renewcommand{\thefigure}{S\arabic{figure}}%
     }

\begin{document}

\begin{frontmatter}

\title{Causality-aware counterfactual confounding adjustment for feature representations learned by deep models}

\runtitle{Counterfactual confounding adjustment for learned representations}

\author{Elias Chaibub Neto \\ Sage Bionetworks, Seattle, WA 98121}

\runauthor{Chaibub Neto E.}

\begin{abstract}
Causal modeling has been recognized as a potential solution to many challenging problems in machine learning (ML). Here, we describe how a recently proposed counterfactual approach developed to deconfound linear structural causal models can still be used to deconfound the feature representations learned by deep neural network (DNN) models. The key insight is that by training an accurate DNN using softmax activation at the classification layer, and then adopting the representation learned by the last layer prior to the output layer as our features, we have that, by construction, the learned features will fit well a (multi-class) logistic regression model, and will be linearly associated with the labels. As a consequence, deconfounding approaches based on simple linear models can be used to deconfound the feature representations learned by DNNs. We validate the proposed methodology using colored versions of the MNIST dataset. Our results illustrate how the approach can effectively combat confounding and improve model stability in the context of dataset shifts generated by selection biases.
\end{abstract}

\end{frontmatter}

\section{Introduction}

Deep neural network (DNN) models are able to learn complex prediction rules from biased data, tarnished by confounding factors and selection biases~\cite{irm2019}. Training DNN models in biased datasets allows the models to carelessly leverage all correlations in the data, including spurious correlations contributed by these data biases, which are unrelated to the causal mechanisms of interest. The unwanted consequence is a potential decrease in generalization performance under dataset shift~\cite{quionero2009}. As a hypothetical example, consider a slight variant of the though experiment described in~\cite{irm2019}, where the goal is to classify cows versus camels. Suppose that the training set is comprised of pictures taken outdoors under natural settings where 99\% of pictures of cows were taken in green pastures while 99\% of pictures of camels are taken in deserts. In this situation it is possible that a convolutional neural network (CNN) might learn to exploit landscape color to classify cows versus camels, so that the CNN might easily misclassify pictures of cows taken, for instance, in a beach. If this is the case, and we want to evaluate the classifier performance on a test set where only 80\% of the cow and camel pictures are taken in green pastures and beige deserts, respectively, then we might see degraded generalization performance.

In situations where the context where the picture is taken (e.g., green pasture versus desert in the above example) is available, we can use this observed confounding variable to try to remove, or at least reduce, the impact of the confounder on the predictive performance of the classifier. Confounding adjustment is an active research area in ML, with balancing approaches being commonly adopted remedies to prevent DNNs from leveraging spurious associations induced by confounders.

\begin{wrapfigure}{r}{0.6\textwidth}
\includegraphics[width=2.85in]{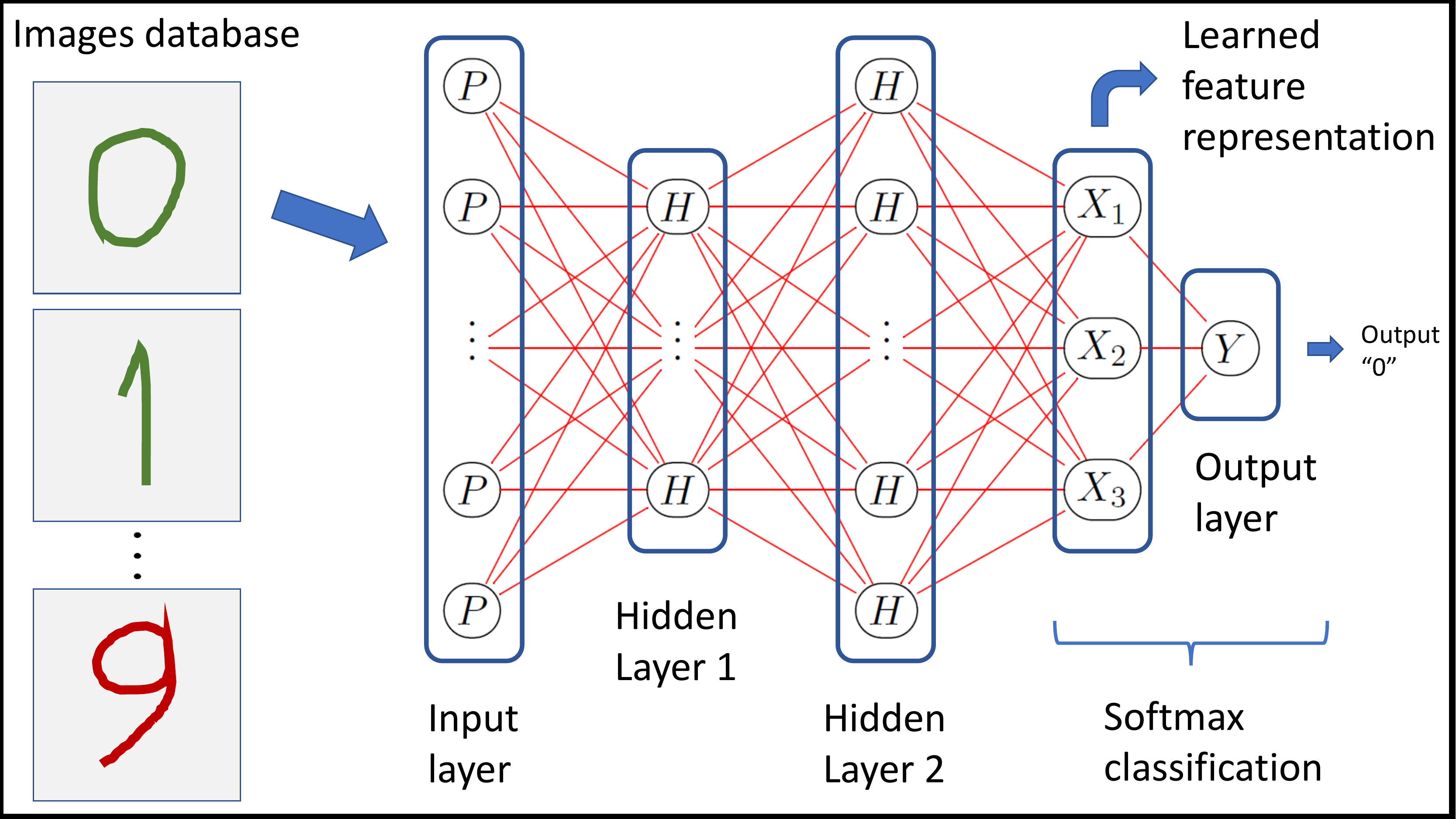}
\caption{The softmax classification step involving the last two layers of this cartoon DNN represent the ``linear modeling step" performed by DNNs.}
\label{fig:linearization.trick}
\vskip - 0.1in
\end{wrapfigure}

In this paper, we describe the applicability of the causality-aware counterfactual confounding adjustment, proposed in~\cite{achaibubneto2020a}, to DNN applications. While the approach in reference~\cite{achaibubneto2020a} is only directly applicable to linear structural causal models, here we show how it can be easily adapted to remove confounding biases from feature representations learned by DNNs. The key insight is that by training a highly accurate DNN using softmax activation at the classification layer, we have that, by construction, the feature representation learned by the last layer prior to the output layer (as illustrated in Figure \ref{fig:linearization.trick}) will fit well a logistic regression model (since the softmax classification used to generate the output of the DNN is essentially performing logistic regression classification). Therefore, even though the associations between the raw inputs and the labels can be highly non-linear, the associations between the learned representation/features and the labels can be well captured by a linear classifier. As a consequence, we can use simple linear models to adjust for confounding. \textit{Throughout the text we denote this process of learning a feature representation that fits well a linear classifier as the ``linear modeling step" performed by DNNs.}

We validate the proposed counterfactual confounding adjustment using a variant of the ``colored MNIST" dataset described in reference~\cite{irm2019}, which corresponds to a synthetic binary classification dataset derived from the MNIST data. Following~\cite{irm2019}, we assign label ``0" to digits $\{0, 1, 2, 3, 4\}$, and label ``1" to digits $\{5, 6, 7, 8, 9 \}$, and color each digit image as either red or green, in a way that most of the training set images labeled as ``0"  are assigned the green color and most of the training set images labeled as ``1" are assigned the red color. Our experiments show that while increasing amounts of dataset shift~\cite{quionero2009} lead to increasing degradation of the generalization performance for classifiers trained without any adjustment, the classifiers built with counterfactually adjusted features still achieved strong generalization performance and stability across all levels of dataset shift. We also show that in the absence of dataset shift, a situation where the color variable improves the predictive performance, the counterfactual adjustment leads to a decrease in predictive performance, as it prevents the classifier from exploiting the image color for the predictions. Comparison against a variation of the SMOTE~\cite{chawla2002} approach (where we balance the joint training distribution of confounders and labels) showed that the causality-aware approach generates predictions that are considerably more stable than the predictions generated by this alternative confounding adjustment approach. This illustrates the point that, while training a learner on unconfounded training data can prevent it from learning the confounding signal (and, therefore, improve model stability), adjusting the training data alone can be insufficient, and better stability can be achieved by approaches, such as the causality-aware method, which also deconfound the test set features.

%In order to validate the causality-aware approach in a more challenging dataset, we also apply the method to real world data collected from a mobile health study in Parkinson's disease (PD)~\cite{bot2016}. In this study, the goal is to classify PD versus non-PD subjects using accelerometer data collected by consumer grade smartphones. Due to selection biases~\cite{heckman1979,hernan2004} during recruitment for this study, age is an important confounder in this dataset~\cite{chaibubneto2019}.

The rest of this paper is organized as follow. Section 2 presents notation and some background. Section 3 describes related work and contrasts the proposed approach against invariant and stable prediction approaches in the literature. Section 4 describes the causal models underlying our illustration. Section 5 describes the causality-aware an SMOTE adjustment approaches. Section 6 presents our illustrative example, and Section 7 describes a simply approach that can be used to empirically check the effectiveness of the counterfactual adjustment. Final remarks are provided in Section 8.

\section{Notation and background}

Throughout the text, we let $Y$, $C$, and $P_{ix}$ represent, respectively, the classification labels, the confounder variable, and the raw data (e.g., pixels in images). We use $X$ to denote a feature representation learned by a DNN, and denote the set of learned features by $\bfX = \{X_1, X_2, \ldots, X_k \}$, where $k$ represents the number of units in the hidden layer prior to the output layer (e.g., $k=3$ in the cartoon DNN in Figure 1). The causality-aware counterfactual versions of the learned features are represented by $X^\ast$. We let the (sub) superscripts $tr$ and $ts$ represent, respectively, the training and test sets. In this paper, we focus on anticausal classification tasks~\cite{scholkopf2012} confounded by selection mechanisms (as described in detail in Section 4). We assume that the causal effects of the confounders on the inputs and on the labels, as well as, the causal effects of the labels on the inputs are stable across the training and test sets, and that dataset shifts arise only due to selection biases. In other words, we assume that the dataset shifts are restricted to the $Pr(\bfC, Y)$, while $Pr(P_{ix} \mid \bfC, Y)$ is stable.

%But, before we describe how to evaluate the causality-aware adjustment, we first provide some additional background needed for these assessments.

Following~\cite{pearl2009,spirtes2000}, we adopt a mechanism-based approach to causation, where the statistical information encoded in the joint probability distribution of a set of random variables is supplemented by a \textit{directed acyclic graph} (DAG) describing the (assumed) causal relations between the variables. In this framework, a \textit{path} in a DAG is defined as any unbroken, nonintersecting sequence of edges in a graph, which may go along or against the direction of the arrows. A path is said to be \textit{d-separated} or \textit{blocked}~\cite{pearl2009} by a set of nodes $\bfW$ if and only if: (i) the path contains a chain $V_j \rightarrow V_m \rightarrow V_k$ or a fork $V_j \leftarrow V_m \rightarrow V_k$ such that the middle node $V_m$ is in $\bfW$; or (ii) the path contains a collider $V_j \rightarrow V_m \leftarrow V_k$ such that $V_m$ is not in $\bfW$ and no descendant of $V_m$ is in $\bfW$. Otherwise, the path is said to be \textit{d-connected} or \textit{open}. A joint probability distribution over a set of variables is \textit{faithful}~\cite{spirtes2000,pearl2009} to a causal diagram if no conditional independence relations, other than the ones implied by the d-separation criterion are present. We adopt the notation $V_1 \nci V_2$ and $V_1 \ci V_2$ to represent marginal statistical dependence and independence, respectively. Conditional dependencies and independencies of $V_1$ and $V_2$ given $V_3$ are represented, respectively, by $V_1 \nci V_2 \mid V_3$ and $V_1 \ci V_2 \mid V_3$.

\section{Related work}

This work describes how the causality-aware method~\cite{achaibubneto2020a}, a recently proposed counterfactual approach developed to deconfound linear structural causal models, can still be used to deconfound the feature representations learned by DNN models. Similarly to invariant prediction approaches~\cite{peters2016,ghassami2017,heinze2018,rojascarulla2018,magliacane2018,irm2019} and stable prediction approaches~\cite{kuang2018,subbaswamy2018,subbaswamy2019,kuang2020}, the goal is to generate predictions based on the stable properties of the data, without absorbing unstable spurious associations. (See~\cite{achaibubneto2020a} for descriptions and contrasting of these approaches against the causality-aware method.)

As clearly articulated by~\cite{subbaswamy2020} there are, broadly speaking, two types of stable prediction approaches: (i) \textit{reactive} methods, that use data (or knowledge) from the intended deployment/target population to correct for shifts; and (ii) \textit{proactive} methods, that do not require data from the deployment/target populations, and are able to learn models that are stable with respect to unknown dataset shifts. Many reactive approaches in the literature~\cite{shimodaira2000,sugiyama2007,dudik2006,huang2007,gretton2009,bickel2009,liu2014} deal with dataset shift by reweighting the training data to make it more closely aligned it with the target test distribution. 

%In this paper, however, we focus on anticausal prediction tasks~\cite{scholkopf2012} and address only dataset shifts in the joint distribution of the confounders and outcome variable, $Pr(\bfC, Y)$, caused by selection biases~\cite{heckman1979,hernan2004,bareinboim2012}.

In situations where the target (test set) joint distribution $Pr(\bfC_{ts}, Y_{ts})$ is known a priori, we can combat dataset shifts by applying simple balancing approaches to the training data (in order to make $Pr(\bfC_{tr}, Y_{tr})$ match $Pr(\bfC_{ts}, Y_{ts})$). In this paper, however, we assume that the dataset shifts in $Pr(\bfC_{ts}, Y_{ts})$ are unknown. In this more challenging situation, it is a common practice to train ML models on deconfounded training data, with the hope that it will prevent the models from learning the confounding signal, so that the learners may show better stability when applied in shifted test sets. (For instance, this is a common strategy to combat discrimination in fairness research, where data pre-processing techniques such as re-weighting and (under-) over-sampling are applied to the training data alone, in order remove the association between sensitive variables and the classifier labels~\cite{calders2009,kamiran2012}). Here, we compare the causality-aware approach against a balancing approach (applied to the training set alone) in order to show that better stability can be achieved by removing confounding from the test set features as well.

\section{Anticausal classification tasks confounded by selection biases}

\subsection{The causal model underlying image classification databases}

As pointed in reference~\cite{irm2019}, in image classification tasks causation does not happen between pixels, but between the real-world concepts captured by the camera. Here, we describe the ``real-world causal mechanisms" generating the observed variables, $Y$, $C$, and $P_{ix}$, in a image classification database. For concreteness, we describe it in terms of the cows versus camels classification example, and let the nature variable $Y = \{\mbox{cow}, \mbox{camel}\}$ represent the labels, and $A = \{\mbox{green}, \mbox{beige}\}$ represent a surrogate variable for the environment (pasture or desert) where these animals live under natural circumstances. We let $S$ represent a binary variable indicating the presence of a selection bias mechanism (described in more detail below).

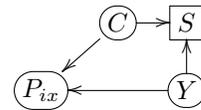
\begin{wrapfigure}{r}{0.22\textwidth}
\vskip -0.1in
$$
\xymatrix@-1.0pc{
& *+[F-:<10pt>]{C} \ar[dl] \ar[r] & *+[F]{S} \\
*+[F-:<10pt>]{P_{ix}}  & & *+[F-:<10pt>]{Y} \ar[ll] \ar[u] \\
}
$$
\vskip -0.1in
  \caption{Causal diagram underlying a image classification database.}
  \label{fig:causal.diagram.Y.A.P}
  \vskip -0.1in
\end{wrapfigure}

Figure \ref{fig:causal.diagram.Y.A.P} presents the causal diagram. The arrow $Y \rightarrow P_{ix}$ indicates that the real world cow or camel causes the observed patterns and intensities of pixels in the image. (Clearly, the observed pixels in the picture cannot cause a cow or a camel in the real-world\footnote{One can argue that looking at the picture causes the human cognition to label it as a cow or a camel. However, this ``human cognition causal process" is different from the ``real-world causal process" generating the observed variables. This distinction between real-world causal mechanisms and human cognition causal processes is clearly presented in Figure 6 of reference~\cite{irm2019}. Because we are interested in modeling the causal processes underlying a machine learning classification task, in our work we will focus on the real-world causal mechanism giving rise to the observed data.}.) Similarly, the arrow $C \rightarrow P_{ix}$ indicates that the green pasture or beige desert environments also influence the observed intensities and patterns of pixels in the image. Finally, note that the arrows pointing from both $Y$ and $C$ towards the square-framed $S$ variable indicates the presence of a selection mechanism generating an association between $Y$ and $C$ in the database. Here, the squared frame around $S$ indicates that we are conditioning on the $S$ variable.

As an example, consider a database (denoted database 1) where the cow pictures were taken in organic livestock farms in France, while the camel pictures were taken in Middle East deserts. For this database, most of the cow pictures will be associated with green pasture backgrounds, while most of the camel pictures will be associated with beige desert backgrounds. Hence, conditional on the fact that database 1 was generated from pictures taken in France and in the Middle East, we observe a strong association between $Y$ and $C$. In this example, we define $S$ as a binary variable assuming value 1 when a picture was taken in France or in the Middle East, and value 0 otherwise. Because any ML models generated from database 1 are conditional on the fact that that the data came from France or the Middle East, i.e., $S = 1$, we represent $S$ using a square frame. (Note that by conditioning on $S = 1$ the $C$ and $Y$ variables become d-connected~\cite{pearl2009}.) Now, consider a second database (denoted database 2) where the cow pictures were taken in natural grassland farms in semi-arid African countries and the camel pictures were taken in the Sahara desert. In this second dataset, the association between the green/beige background color and the cow/camel labels is expected to be weaker than in database 1, because except for the cow pictures taken during the rainy season, when the natural grasslands are green, the pictures taken during the dry season will also show beige backgrounds of dried grasslands. For this second dataset, we have that $S$ represents a binary variable assuming value 1 when a picture was taken in Africa, and 0 otherwise. Due to the different selection mechanisms, there is a clear shift in the joint distribution of $\{C, Y\}$ in database 2 relative to database 1.

\section{Confounding adjustments}

\begin{wrapfigure}{r}{0.45\textwidth}
\vskip -0.2in
$$
\xymatrix@-1.1pc{
(a) & *+[F-:<10pt>]{C} \ar[dl] & & (b) & *+[F-:<10pt>]{C} \ar[r] & *+[F]{S} \\
*+[F-:<10pt>]{P_{ix}}  & & *+[F-:<10pt>]{Y} \ar[ll] & *+[F-:<10pt>]{P_{ix}^\ast}  & & *+[F-:<10pt>]{Y} \ar[ll] \ar[u] \\
}
$$
\vskip -0.1in
  \caption{Confounding adjustments.}
  \label{fig:conf.adjusts}
  %\vskip -0.1in
\end{wrapfigure}
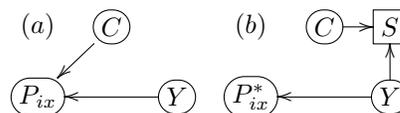

The causal diagram in Figure \ref{fig:causal.diagram.Y.A.P} shows that $C$ represents a confounder of the relationship between the labels, $Y$, and $P_{ix}$. Figure \ref{fig:conf.adjusts} describes two ways we can potentially remove the spurious associations between $Y$ and $P_{ix}$ generated by $C$. Panel a shows that we could attempt to remove the association between $Y$ and $C$ by balancing the data, while panel b shows that we could try to remove the causal influence of $C$ on $P_{ix}$ by using a counterfactual approach. In this paper, we take the latter route. However, instead of attempting to directly model
the highly non-linear causal relations $Y \rightarrow P_{ix}$ and $C \rightarrow P_{ix}$, we leverage the ``linear modeling step" described before and work with the learned representation, $\bfX$. Furthermore, we compare the counterfactual approach against a balancing approach (described in subsection 5.2), and illustrate the better stability enjoyed by the counterfactual approach under dataset shifts of the joint distribution $Pr(\bfC, Y)$.

\subsection{Causality-aware counterfactual confounding adjustment for learned representations}

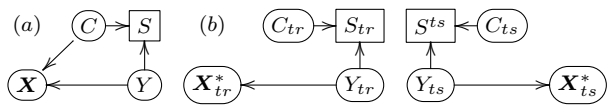
\begin{wrapfigure}{r}{0.67\textwidth}
\vskip -0.1in
{\footnotesize
$$
\xymatrix@-1.3pc{
(a) & *+[F-:<10pt>]{C} \ar[dl] \ar[r] & *+[F]{S} & (b) & *+[F-:<10pt>]{C_{tr}} \ar[r] & *+[F]{S_{tr}} & *+[F]{S^{ts}} & *+[F-:<10pt>]{C_{ts}} \ar[l] & \\
*+[F-:<10pt>]{\bfX}  & & *+[F-:<10pt>]{Y} \ar[ll] \ar[u] & *+[F-:<10pt>]{{\bfX_{tr}^\ast}}  & & *+[F-:<10pt>]{Y_{tr}} \ar[ll] \ar[u] & *+[F-:<10pt>]{Y_{ts}} \ar[rr] \ar[u] & & *+[F-:<10pt>]{{\bfX_{ts}^\ast}} \\
}
$$}
\vskip -0.1in
  \caption{Causality-aware adjustment on learned features.}
  \label{fig:learned.features.causal.diagrams}
  \vskip -0.1in
\end{wrapfigure}
Using the ``linear modeling step" we generate a set of learned features, $\bfX$. Note that training a logistic regression classifier using $\bfX$ as inputs is essentially equivalent as training a DNN on the raw input data. Because $\bfX$ corresponds to a transformation of the raw data $P_{ix}$, we can still represent the data generation process associated with $Y$, $C$, and $\bfX$ by the causal diagram in Figure \ref{fig:learned.features.causal.diagrams}a, where $P_{ix}$ is replaced by $\bfX$.

However, training a logistic regression classifier on $\bfX$ will still generate predictions biased by $C$. In order to remove/reduce the influence of $C$ on the predictive performance of the classifier, we apply the causality-aware adjustment proposed by~\cite{achaibubneto2020a} to generate counterfactual features, ${\bfX^\ast}$, according to the causal diagram in Figure \ref{fig:learned.features.causal.diagrams}b. The approach is implemented according to Algorithm 1:
\begin{algorithm}[!h]
\caption{Causality-aware learned feature representations}\label{alg:counterfactual.adjustment}
\KwData{Training image data, $P_{ix}^{tr}$; training confounder data, $\bfC_{tr}$, training label data $Y_{tr}$; test-set image data, $P_{ix}^{ts}$; test-set confounder data, $\bfC_{ts}$.}
\ShowLn Fit a DNN to the training images. \\
\ShowLn Propagate the raw data from each image in the training and test set through the DNN and extract the respective learned features (which correspond to the values generated by the hidden-layer of the DNN closer to the output layer). \\
\ShowLn \For{each learned feature $X_j$} {
\ShowLn $\bullet$ Using the training set, estimate regression coefficients and residuals from the linear regression model, $X_j^{tr} = \mu_j^{tr} + \beta_{{X_j} Y}^{tr} \, Y_{tr} + \sum_i \beta_{{X_j} {C_i}}^{tr} \, C^{tr}_i + W_{X_j}^{tr}$, and then compute the respective counterfactual feature, $\hat{X}_j^{\ast tr} = X_j^{tr} - \sum_i \hat{\beta}_{{X_j} {C_i}}^{tr} \, C^{tr}_i = \hat{\mu}_j^{tr} + \hat{\beta}_{{X_j} Y}^{tr} \, Y_{tr} + \hat{W}_{X_j}^{tr}$~. \\
\ShowLn $\bullet$ Using the test set, compute the counterfactual feature, $\hat{X}_j^{\ast ts} = X_j^{ts} - \sum_i \hat{\beta}_{{X_j} {C_i}}^{tr} \, C^{ts}_i$~.
}
\KwResult{Counterfactual training and test features, $\hat{\bfX}^{\ast}_{tr}$ and $\hat{\bfX}^{\ast}_{ts}$.}
\end{algorithm}

Once the counterfactual features have been generated by Algorithm 1, we can then use $\hat{\bfX}^{\ast}_{tr}$ to train a logistic regression classifier, and then use $\hat{\bfX}^{\ast}_{ts}$ to generate predictions that are no longer biased by the confounder (or, at least, impacted by a lesser degree). Note that we do not make use of the test set labels in the estimation of the counterfactual test set features, and that step 5 of Algorithm 1 uses the regression coefficients estimated in the training set, $\hat{\beta}_{{X_j} {C_i}}^{tr}$, for the computation of the test set counterfactual features. This, however, assumes that $Pr(P_{ix}^{tr} \mid Y_{tr}, \bfC_{tr}) = Pr(P_{ix}^{ts} \mid Y_{ts}, \bfC_{ts})$\footnote{Note that if $Pr(P_{ix}^{tr} \mid Y_{tr}, \bfC_{tr}) = Pr(P_{ix}^{ts} \mid Y_{ts}, \bfC_{ts})$ holds, then we have that $Pr(X_j^{tr} \mid Y_{tr}, \bfC_{tr}) = Pr(X_j^{ts} \mid Y_{ts}, \bfC_{ts})$ will also hold, since both the training and test features $\hat{\bfX}_{tr}$ and $\hat{\bfX}_{ts}$ are generated by propagating the raw training and test data, $P_{ix}^{tr}$ and $P_{ix}^{ts}$, on the same trained DNN. This, by its turn, implies that the asymptotic estimates of the regression coefficients $\mu_j$, $\beta_{{X_j} Y}$, and $\beta_{{X_j}{C_i}}$ will be similar in the training and test sets, so that, for large enough sample sizes (what is usually the case in deep learning applications), we have that the estimation of the causality-aware test features using, $X_j^{\ast ts} = X_j^{ts} - \sum_i \hat{\beta}_{{X_j} {C_i}}^{tr} \, C^{ts}_i$, is equivalent to estimate $X_j^{\ast ts}$ using $X_j^{\ast ts} = \hat{\mu}_j^{ts} + \hat{\beta}_{{X_j} Y}^{ts} \, Y_{ts} + \hat{W}_{X_j}^{ts}$ since, $X_j^{ts} - \sum_i \hat{\beta}_{{X_j} {C_i}}^{tr} \, C^{ts}_i \approx X_j^{ts} - \sum_i \hat{\beta}_{{X_j} {C_i}}^{ts} \, C^{ts}_i = \hat{\mu}_j^{ts} + \hat{\beta}_{{X_j} Y}^{ts} \, Y_{ts} + \hat{W}_{X_j}^{ts}$.}.

\subsection{SMOTE balancing adjustment}

We compared the proposed counterfactual confounding method against a variant of the SMOTE (Synthetic Minority Over-sampling TEchnique) adjustment~\cite{chawla2002}. Similarly to SMOTE, where the goal is to balance the label classes by under-sampling examples from the majority class and over-sampling synthetically modified examples from the minority class, our variation under-samples from the label/color majority categories and over-samples synthetically modified examples from the minority categories, in order to balance the label/confounder categories. We denote this variation of SMOTE as ``SMOTE balancing". Note that this approach is used to improve the balance of the joint distribution of the confounders and labels, $Pr(\bfC, Y)$, in the training set (but not on the test sets).

\section{Illustrations}

\subsection{Colored MNIST experiments}

\begin{figure}[!h]
%\vskip -0.5in
\includegraphics[width=\linewidth]{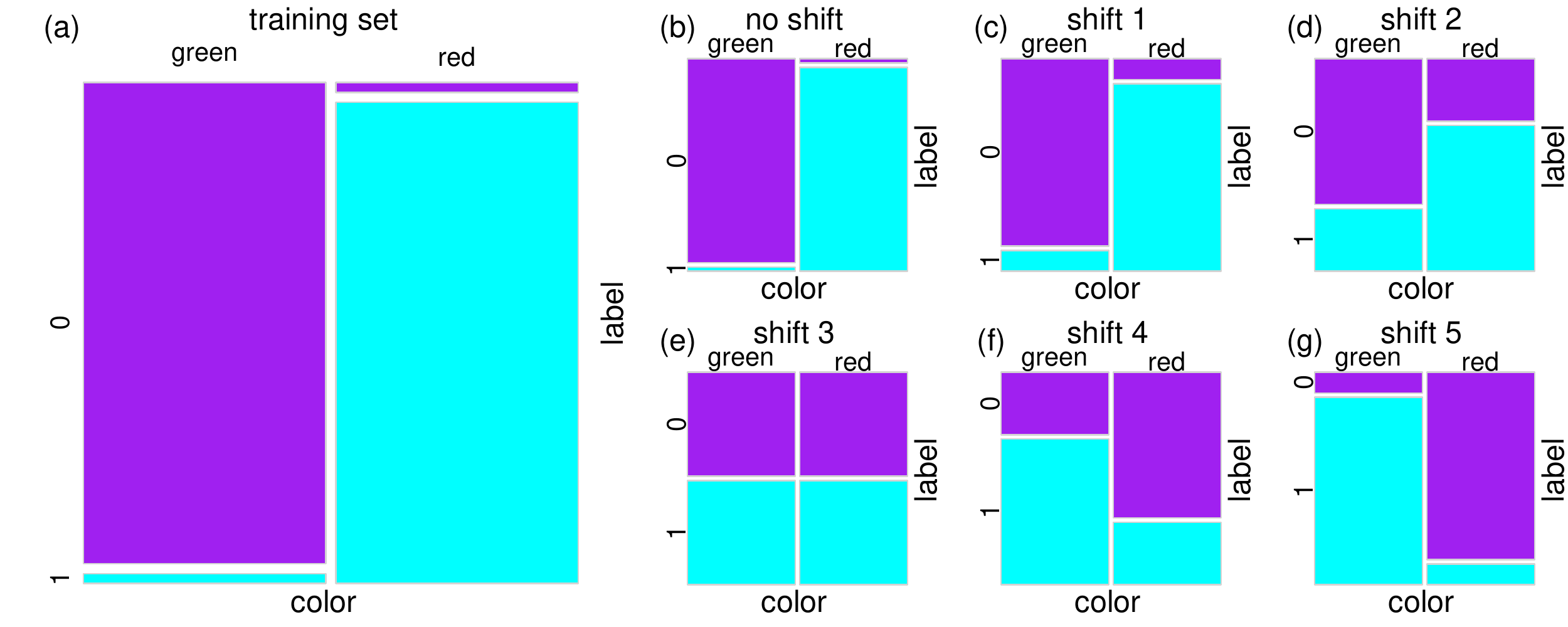}
\vskip -0.1in
\caption{Joint distributions of the label and color variables in the training set (panel a) versus 6 distinct test sets in the absence (panel b) and presence (panels c, d, e, f, and g) of dataset shift.}
\label{fig:shift.mosaic.plots}
%\vskip -0.1in
\end{figure}

We validate the metholodology using a variant of the ``colored MNIST" dataset described in~\cite{irm2019}, which corresponds to a synthetic binary classification dataset derived from the MNIST data. Following reference~\cite{irm2019}, we assign label ``0" to digits $\{0, 1, 2, 3, 4\}$, and label ``1" to digits $\{5, 6, 7, 8, 9 \}$, and color each digit image as either red or green, in a way that most of the training set images labeled as ``0"  are assigned the green color and most of the training set images labeled as ``1" are assigned the red color. Using this dataset we evaluate the effectiveness of the approach in a series of experiments in both the: (i) absence of dataset shift (``no shift" experiment), where the test set is colored in exactly the same manner as the training set; and (ii) presence of increasing amounts of dataset shift (``shift 1" to ``shift 5" experiments), where we decrease the proportion of green images labeled as ``0" and red images labeled as ``1" in the test set, in order to generate a shift in the joint distribution of the color and label variables, $Pr(C, Y)$, relative to the training set, as described in Figure \ref{fig:shift.mosaic.plots}.

Note that, in these experiments, digit labels represent stable properties while digit colors represent unstable properties which vary from training to test set, so that the digit color simulates the effect of a confounder\footnote{Strictly speaking, we colored the data in a way that color actually represents a mediator, rather than a confounder. This was done, nonetheless, for computational convenience. We clarify, however, that this has no impact in the validity of our results because, in anticausal prediction tasks, the causality-aware approach handles confounders and mediators in exactly the same way. See Supplementary Section 1.1 for further details.}. In all experiments we trained convolutional neural network (CNN) models (see Supplementary Section 1 for a description of CNN architectures, optimization parameter choices, and other experimental details). We compared the causality-aware approach against the SMOTE balancing approach, where we trained CNNs using balanced data\footnote{In our experiments, the training set (Figure \ref{fig:shift.mosaic.plots}a) contained the following four $\{\mbox{label}, \mbox{color}\}$ categories (at the following proportions): $\{``0", ``\mbox{green}"\}$ (49\%); $\{``1", ``\mbox{green}"\}$ (1\%); $\{``0", ``\mbox{red}"\}$ (1\%); and $\{``1", ``\mbox{red}"\}$ (49\%). Application of the SMOTE balancing generated datasets with 25\% of examples in each of the 4 categories where we randomly under-sampled images from the $\{``0", ``\mbox{green}"\}$ and $\{``1", ``\mbox{red}"\}$ categories, and over-sampled synthetically modified versions of images in the $\{``0", ``\mbox{red}"\}$ and $\{``1", ``\mbox{green}"\}$ categories. (Namely, for each one of the 600 images in each of these minority categories we generated 24 additional randomly rotated versions of the original image.)}, as well as, against the ``no adjustment" approach where we train and evaluate the CNNs on biased data.

\begin{wrapfigure}{r}{0.4\textwidth}
%\vskip -0.1in
\includegraphics[width=2in]{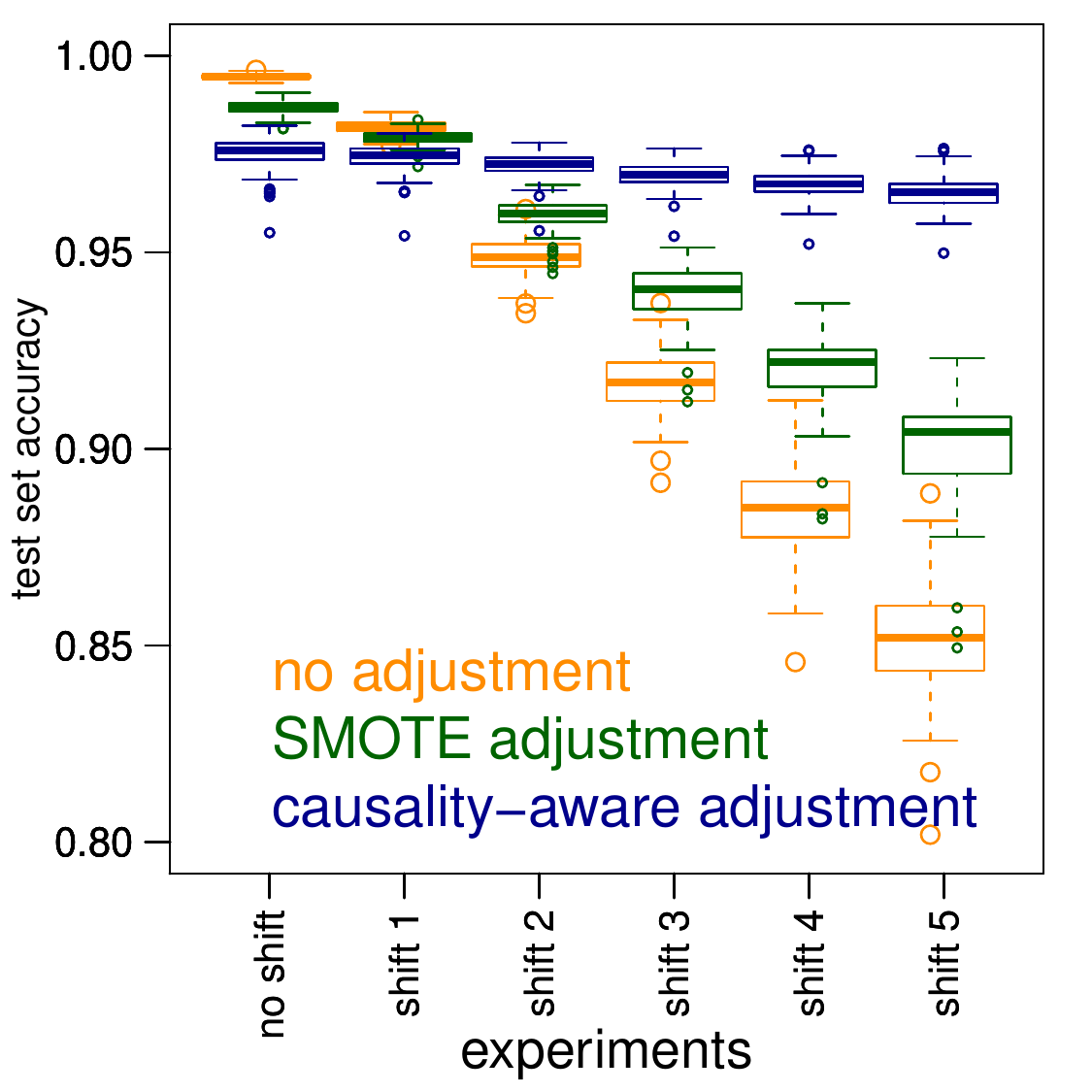}
%\vskip -0.1in
\caption{Colored MNIST experiments.}
\label{fig:colored.mnist.results}
%\vskip -0.1in
\end{wrapfigure}

Figure \ref{fig:colored.mnist.results} reports the results of 100 distinct replications of our experiment based on distinct colorings of the training and test set images. As expected, for classifiers trained without any adjustment (orange boxplots), our experiments show that increasing amounts of dataset shift (described in Figure \ref{fig:shift.mosaic.plots}) lead to increasing degradation of the generalization performance. The classifiers built with counterfactually adjusted features (blue boxplots), on the other hand, achieved strong generalization performance across all levels of dataset shift. 
The results also show that in the absence of dataset shift (or under slightly shift), a situation where the color variable improves the predictive performance, the counterfactual adjustment leads to a decrease in predictive performance, as it prevents the classifier from exploiting the image color for the predictions (compare the orange and blue boxplots in the ``no shift" and ``shift 1" experiments). The green boxplots report the results based on SMOTE balancing, and shows that while the balancing adjustment improved generalization relative to the ``no adjustment" experiments it showed considerably more degradation in performance compared to the causality-aware adjustment under stronger dataset shifts.

As described in~\cite{achaibubneto2020a} the better stability of the causality-aware approach is explained by the fact that the adjustment deconfounds both the training and test set features. To see why, note that the classification accuracies reported in Figure \ref{fig:colored.mnist.results} were computed using logistic regression models fitted to the feature representations learned by the CNNs. Hence, in the case of ``no adjustment" the expected classification accuracy is given by,
\begin{align}
E[\ind\{ Y_{ts} = \hat{Y} \}] &= Pr(Y_{ts} = \hat{Y}) \nonumber \\
&= Pr(Y_{ts} = g(\bfX_{ts} \hat{\bfBeta}_{tr})) \nonumber \\
&= Pr(Y_{ts} = g(\textstyle \sum_j \hat{\beta}_j^{tr} X_j^{ts})) \nonumber \\
&= Pr(Y_{ts} = g(\textstyle \sum_j \hat{\beta}_j^{tr} f(Y_{ts}, C_{ts}, U_{X_j}^{ts}))) \label{eq:expected.accuracy.1}
\end{align}
where: $\hat{\bfBeta}_{tr}$ represents the weights learned by the logistic regression classifier estimated on the confounded training data; the structural causal model $X_j^{ts} = f(Y_{ts}, C_{ts}, U_{X_j}^{ts})$ indicates that $X_j$ is a function of $Y$ and $C$ (and $U_{X_j}$); and $g()$ represents the indicator function,
\begin{equation*}
g(\bfX_{ts} \hat{\bfBeta}_{tr}) = \ind\left\{[1/(1 + \exp{\{- \bfX_{ts} \hat{\bfBeta}_{tr}\}})] \geq t \right\}~,
\end{equation*}
assuming value 1 when the logistic function is greater or equal to the probability threshold $t$, and 0 otherwise. (In our experiments, we set $t = 0.5$.) Because the expression in equation (\ref{eq:expected.accuracy.1}) is still a function of $Y_{ts}$ and $C_{ts}$, it will be unstable with respect to shifts in $Pr(C_{ts}, Y_{ts})$.

Similarly, the expected accuracy for the SMOTE balancing approach is given by,
\begin{align*}
E[\ind\{ Y_{ts} = \hat{Y}^{S} \}] &= Pr(Y_{ts} = g(\textstyle \sum_j \hat{\beta}_j^{tr,S} X_j^{ts})) \\
&= Pr(Y_{ts} = g(\textstyle \sum_j \hat{\beta}_j^{tr,S} f(Y_{ts}, C_{ts}, U_{X_j}^{ts})))~,
\end{align*}
where $\hat{\beta}_j^{tr,S}$ represents the weights learned by the logistic model trained on the balanced training data. Note that even though the weights are estimated in balanced training data (and, therefore, do not absorb spurious associations generated by the confounder) the expected accuracy is still a function of $Y_{ts}$ and $C_{ts}$ (since $X_j^{ts} = f(Y_{ts}, C_{ts}, U_{X_j}^{ts})$), and will still be influenced by shifts in the joint distribution $Pr(C_{ts}, Y_{ts})$.

The expected accuracy of the causality-aware approach, on the other hand, is given by,
\begin{align*}
E[\ind\{ Y_{ts} = \hat{Y}^\ast \}] &= Pr(Y_{ts} = g(\textstyle \sum_j \hat{\beta}_j^{\ast tr} X_j^{\ast ts})) \\
&= Pr(Y_{ts} = g(\textstyle \sum_j \hat{\beta}_j^{\ast tr} f^\ast(Y_{ts}, U_{X_j}^{ts})))~,
\end{align*}
where $\hat{\beta}_j^{\ast tr}$ represents the weights learned by the logistic regression model trained with deconfounded training features. Now, because the approach also deconfounds the test set features (so that $X_{j}^{\ast ts} = f^\ast(Y_{ts}, U_{X_j}^{ts})$ is no longer a function of $C_{ts}$), we have that the expected accuracy no longer depends on $C_{ts}$\footnote{Or, at least, depends on $C_{ts}$ to a lesser degree since in situations where, in practice, the adjustment does not work perfectly, we may still see some dependence on $C_{ts}$.}, explaining the better stability of the causality-aware approach.

These observations illustrate that, while training a learner on unconfounded training data will prevent it from learning the confounding signal and, therefore, improve model stability, adjusting the training data alone can be insufficient, and better stability can be achieved by deconfounding the test set features as well. 

%Finally, because the above experiments are perhaps too ``synthetic" (in the sense that the colored MNIST dataset is not representative of data from more realistic scenarios), we present next an additional example based on real world data in a considerably more challenging task.

\section{Empirical evaluation of the adjustment's effectiveness}

\begin{figure}[!h]
{\footnotesize
$$
\xymatrix@-0.5pc{
*+[F]{S_{tr}} & *+[F-:<10pt>]{C_{tr}} \ar[l] \ar[d] & (a) & *+[F-:<10pt>]{C_{ts}} \ar[r] \ar[d] & *+[F]{S_{ts}} & (b) & *+[F-:<10pt>]{C_{ts}} \ar[r] \ar[dl] & *+[F]{S_{ts}} \\
*+[F-:<10pt>]{Y_{tr}} \ar[u] \ar[r] \ar@/^1.25pc/@[red][rr] & *+[F-:<10pt>]{{\bfX}_{tr}} \ar@[red][r] & *+[F-:<10pt>]{\hat{R}_{ts}} & *+[F-:<10pt>]{{\bfX}_{ts}} \ar@[red][l] & *+[F-:<10pt>]{Y_{ts}} \ar[u] \ar[l] & *+[F-:<10pt>]{\hat{R}_{ts}} & & *+[F-:<10pt>]{Y_{ts}} \ar[u] \ar@[red][ll] \\
*+[F]{S_{tr}} & *+[F-:<10pt>]{C_{tr}} \ar[l] & (c) & *+[F-:<10pt>]{C_{ts}} \ar[r] & *+[F]{S_{ts}} & (d) & *+[F-:<10pt>]{C_{ts}} \ar[r] & *+[F]{S_{ts}} \\
*+[F-:<10pt>]{Y_{tr}} \ar[u] \ar[r] \ar@/^1.25pc/@[red][rr] & *+[F-:<10pt>]{\bfX^\ast_{tr}} \ar@[red][r] & *+[F-:<10pt>]{\hat{R}^\ast_{ts}} & *+[F-:<10pt>]{\bfX^\ast_{ts}} \ar@[red][l] & *+[F-:<10pt>]{Y_{ts}} \ar[u] \ar[l] & *+[F-:<10pt>]{\hat{R}^\ast_{ts}} & & *+[F-:<10pt>]{Y_{ts}} \ar[u] \ar@[red][ll] \\
}
$$}
%\vskip -0.1in
  \caption{Black arrows represent the data generation process of the observed data. Red arrows represent the data generation process of the predictions, $\hat{R}_{ts}$, which correspond to the predicted positive class probability of the test set examples. Note that $\hat{R}_{ts}$ is a function of the training data (used to train the classifier), and the test set inputs (used to generate the predictions).}
  \label{fig:classification.causal.diagram}
  %\vskip -0.1in
\end{figure}
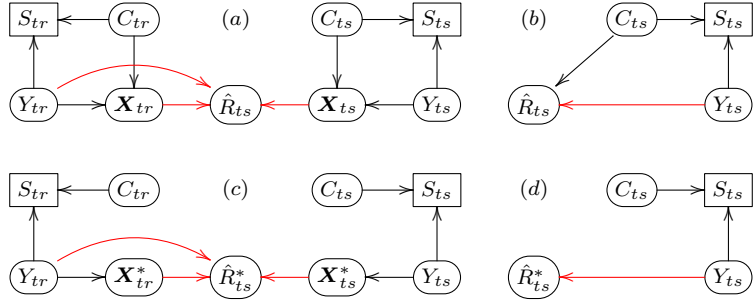

Here, we describe how we can empirically evaluate if the causality-aware adjustment is effectively deconfounding the predictions. In practice, it is important to perform these ``sanity checks" in order to make sure that the linear models used to deconfound the learned feature representations are really able to remove the direct confounder effects from the counterfactual features. Following~\cite{chaibubneto2019}, we describe a simple approach to evaluate if the causality-aware adjustment is indeed removing the spurious associations generated by the confounders from the predictions generated by the logistic regression classifier. The key idea is to represent the data generation process of the observed data together with the data generation process giving rise to the predictions as a causal diagram, and compare the conditional independence relations predicted by d-separation against the conditional independence (CI) relations observed in the data. For instance, Figure \ref{fig:classification.causal.diagram}a represents the full causal diagram for the image classification task, where $\hat{R}_{ts}$ represents the predicted positive class probability of the test set examples. Figure \ref{fig:classification.causal.diagram}b represents a simplified version that displays only the test data and omits the learned representation node, ${\bfX}_{ts}$. Note that for this diagram, $C$ represents a confounder\footnote{Following Pearl~\cite{pearl2009}, we adopt a graphical definition of confounding where a variable $C$ is a confounder of the relationship between variables $X$ and $Y$, if there is an open path from $C$ to $X$ that does not go thorough $Y$, and there is an open path from $C$ to $Y$ that does not go thorough $X$, in the DAG representing the data generation process giving rise to these variables.} of the prediction $\hat{R}_{ts}$, since there is an open path from $C_{ts}$ to $\hat{R}_{ts}$ that does not go through $Y_{ts}$ (namely, $C_{ts} \rightarrow \hat{R}_{ts}$), as well as, an open path from $C_{ts}$ to $Y_{ts}$ that does not go through $\hat{R}_{ts}$ (namely, $C_{ts} \rightarrow S_{ts} \leftarrow Y_{ts}$, where $S_{ts}$ is conditioned on.). Figures \ref{fig:classification.causal.diagram}c and d represent causal diagrams for tasks trained with the causality-aware counterfactual features. Note that in Figure \ref{fig:classification.causal.diagram}d, $C_{ts}$ is no longer a confounder of the predictions $\hat{R}_{ts}$, since the only open path connecting $C_{ts}$ to $\hat{R}_{ts}$ goes through $Y_{ts}$.

If the causality-aware counterfactual adjustment is effective (and faithfulness holds) then, from the application of d-separation~\cite{pearl2009} to the simplified causal graph in Figure \ref{fig:classification.causal.diagram}d, we would expect to see the following pattern of CI relations in the data:
\begin{equation}
\left.
\begin{tabular}{c}
${\hat{R}^\ast_{ts}} \nci Y_{ts}$ \\
${\hat{R}^\ast_{ts}} \nci C_{ts}$ \\
$C_{ts} \nci Y_{ts}$ \\
${\hat{R}^\ast_{ts}} \nci Y_{ts} \mid C_{ts}$ \\
${\hat{R}^\ast_{ts}} \ci C_{ts} \mid Y_{ts}$ \\
$C_{ts} \nci Y_{ts} \mid \hat{R}^\ast_{ts}$
\end{tabular}
\right\}
\label{eq:ci.patterns.supple}
\end{equation}
That is, we would expect ${\hat{R}^\ast_{ts}}$ and $C_{ts}$ to be conditionally independent given $Y_{ts}$, since conditioning on $Y_{ts}$ blocks the path $C_{ts} \rightarrow S_{ts} \leftarrow Y_{ts} \rightarrow \hat{R}^\ast_{ts}$ in Figure \ref{fig:classification.causal.diagram}d. On the other hand, if the adjustment has failed (or if no adjustment was performed), we would expect to see the conditional association ${\hat{R}^\ast_{ts}} \nci C_{ts} \mid Y_{ts}$ in the data.

\begin{figure}[!h]
\centerline{\includegraphics[width=\linewidth]{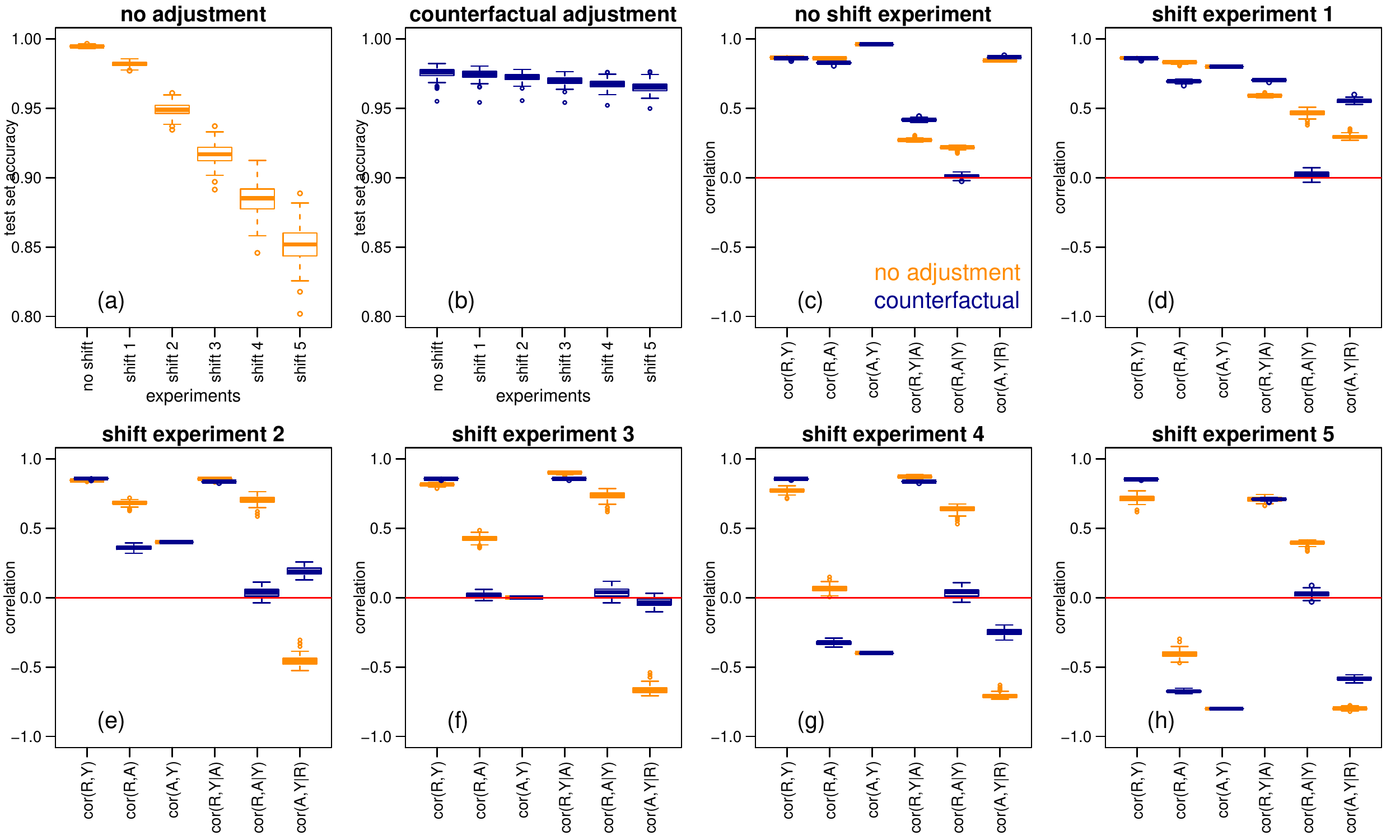}}
\caption{Conditional independence patterns for the colored MNIST experiments.}
\label{fig:colored.mnist.experiments}
\end{figure}

Figure \ref{fig:colored.mnist.experiments} compares the causality-aware approach (blue) against the ``no adjustment" approach (orange) and reports these conditional independence patterns (computed using marginal and partial correlations) for the colored MNIST experiments. Note that, for the causality-aware approach (blue boxplots), in all experiments (panels c to h) we observe that the distribution of the estimated partial correlations $cor(R, C \mid Y)$ is tightly concentrated around zero, suggesting that the adjustment was effective because ${\hat{R}^\ast_{ts}}$ appears to be (linearly) independent of $C_{ts}$ when we condition on $Y_{ts}$. Observe, as well, that this is not the case for the ``no adjustment" approach (orange boxplots), where these partial correlations are located far away from zero.

\begin{figure}[!h]
{\footnotesize
$$
\xymatrix@-0.5pc{
& *+[F-:<10pt>]{C_{tr}} \ar[d] & (a) & *+[F-:<10pt>]{C_{ts}} \ar[d] & & (b) & *+[F-:<10pt>]{C_{ts}} &  \\
*+[F-:<10pt>]{Y_{tr}} \ar[ur] \ar[r] \ar@/^1.25pc/@[red][rr] & *+[F-:<10pt>]{\bfX_{tr}} \ar@[red][r] & *+[F-:<10pt>]{\hat{R}_{ts}} & *+[F-:<10pt>]{\bfX_{ts}} \ar@[red][l] & *+[F-:<10pt>]{Y_{ts}} \ar[l] \ar[ul] & *+[F-:<10pt>]{\hat{R}_{ts}} & & *+[F-:<10pt>]{Y_{ts}} \ar[ul] \ar@[red][ll] \\
& *+[F-:<10pt>]{C_{tr}} & (c) & *+[F-:<10pt>]{C_{ts}} & & (d) & *+[F-:<10pt>]{C_{ts}} &  \\
*+[F-:<10pt>]{Y_{tr}} \ar[ur] \ar[r] \ar@/^1.25pc/@[red][rr] & *+[F-:<10pt>]{\bfX^\ast_{tr}} \ar@[red][r] & *+[F-:<10pt>]{\hat{R}^\ast_{ts}} & *+[F-:<10pt>]{\bfX^\ast_{ts}} \ar@[red][l] & *+[F-:<10pt>]{Y_{ts}} \ar[l] \ar[ul] & *+[F-:<10pt>]{\hat{R}^\ast_{ts}} & & *+[F-:<10pt>]{Y_{ts}} \ar[ul] \ar@[red][ll] \\
}
$$}
%\vskip -0.1in
  \caption{Prediction task data generation process for mediation problems.}
  \label{fig:classification.causal.diagram.2.med}
  %\vskip -0.1in
\end{figure}
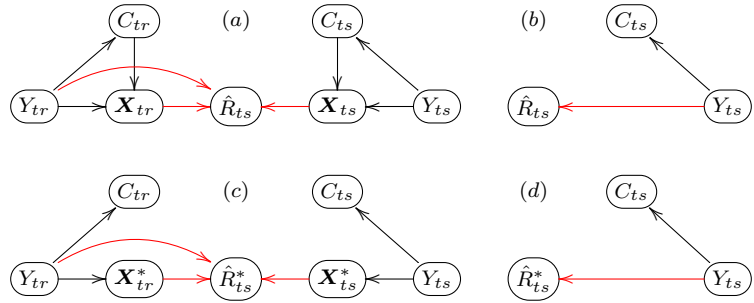

Note that while Figure \ref{fig:classification.causal.diagram} focus on the image classification application, the same set of CI relations in equation \ref{eq:ci.patterns.supple} still hold for the mediation problems, as illustrated in Figure \ref{fig:classification.causal.diagram.2.med}.

%Analogous graphs for mediation problems. Note that these diagrams imply the same CI relations between ${\hat{R}^\ast_{ts}}$, $C_{ts}$, and $Y_{ts}$ as panel d in Figure \ref{fig:classification.causal.diagram}.

%But, before we describe how to evaluate the causality-aware adjustment, we first provide some additional background needed for these assessments.
%This is important, in practice, since the approach depends in parametric modeling choices.
%Here, we describe how it is possible to empirically evaluate if the causality-aware adjustment is really removing the confounding signal from the predictions. This is important, in practice, since the approach depends in parametric modeling choices.

\section{Discussion}

In real life applications, the effective use of counterfactual adjustments often require the specification of a parametric model that fits well the observed data. For instance, attempting to fit a linear regression model to variables showing non-linear associations may lead to the estimation of regression coefficients that fail to capture the associations between outputs and inputs, so that these associations might still be leaked to the residuals. The main insight in this paper is the realization that the ``linear modeling step" performed by DNNs allows for effective counterfactual adjustment in anticausal classification tasks where we can train an accurate DNN. Contrary to tabular data, whose data generation process is completely outside the control of the modeler, the feature representations learned by accurate DNNs are, by construction, shaped according to the adopted neural network architecture. In a sense, the can think of the feature representation learned by a DNN as the output of a data transformation process. We believe that this alternative view of DNNs, as an effective approach to transform raw data into learned representations that fit well a specified parametric model, will make the use of parametric counterfactual approaches more practical and effective.

Finally, the approach has a few additional practical advantages. First, the approach is trivial to implement (as the generation of the counterfactual features only require the computation of linear regression fits). Second, it does not require re-training DNNs and can be applied to deconfound the representations learned by standard DNNs (this can be an important practical advantage in applications involving large neural networks trained on large datasets). Third, the effectiveness of the causality-aware approach can be evaluated empirically using the sanity checks described in Section 7.

\beginsupplement

\vskip 0.5in

\noindent {\Huge Supplement}

\setcounter{section}{0}

\section{Colored MNIST illustration experimental details}

In all colored MNIST experiments we trained CNN models for 10 epochs using the RMSprop optimizer with learning rate 0.001, and mini-batches of 128 images, according with the following architecture: Input $\rightarrow$ Conv2d(32, (3, 3), relu) $\rightarrow$ Conv2d(32, (3, 3), relu) $\rightarrow$ MaxPooling2d((2, 2)) $\rightarrow$ Dropout(0.25) $\rightarrow$ Conv2d(32, (3, 3), relu) $\rightarrow$ Conv2d(32, (3, 3), relu) $\rightarrow$ MaxPooling2d((2, 2)) $\rightarrow$ Dropout(0.25) $\rightarrow$ Dense(16, relu) $\rightarrow$ Dropout(0.5) $\rightarrow$ Dense(2, softmax), where: Conv2d($f$, $s$, $a$) represents a convolutional layer with number of filters given by $f$, strides given by $s$, and activation given by $a$; MaxPooling2d($p$) represents the max-pooling operation with pool-size $p$; Dropout($r$) represents a dropout layer with dropout rate $r$; and Dense($u$, $a$) represents a dense layer with number of units given by $u$, and activation function given by $a$. (Note that the learned representation used in these illustrations corresponds to the values extracted from the 16 units in the Dense(16, relu) layer.) All experiments were performed using the \texttt{keras}~\cite{rkeras2017} R package~\cite{rproject2019} (a R wrapper of Keras\cite{keras2015}). (Alternative training settings produced similar results.)

Figure \ref{fig:colored.mnist.results} in the main text reports the results from 100 distinct replications of the experiment. For each replication, we:
\begin{enumerate}
\item Generated a colored training set and 6 distinct colored test sets as illustrated in Figure \ref{fig:shift.mosaic.plots} in the main text (and described in more detail in Section 1.1 below).
\item Trained a CNN model using the training images, and then extracted the learned training and test feature representations, $\bfX_{tr}$ and $\bfX_{ts,k}$, $k = 1, \ldots, 6$, by propagating the images from the training set and of the 6 distinct test sets on the trained CNN.
\item Trained a CNN model using the SMOTE balanced training images, and extracted the learned feature representations, $\bfX_{tr}^S$ and $\bfX_{ts,k}^S$, $k = 1, \ldots, 6$, by propagating the training and test set images on the SMOTE balanced CNN model. (Note that we used the same exact 6 distinct image test sets as in step 2.)
\item For the ``no adjustment" approach we: trained a logistic regression model using $\{\bfX_{tr}, Y_{tr}\}$; generated the predictions, $\hat{Y}_k$, using $\bfX_{ts,k}$, $k = 1, \ldots, 6$; and evaluated the respective classification accuracies across the 6 test sets.
\item For the SMOTE balancing approach we: trained a logistic regression model using $\{\bfX_{tr}^S, Y_{tr}\}$; generated the predictions, $\hat{Y}_k^S$, using $\bfX_{ts,k}^S$, $k = 1, \ldots, 6$; and then evaluated the respective classification accuracies across the 6 test sets.
\item For causality-aware approach we: computed the counterfactual training and test set features, $\bfX_{tr}^\ast$ and $\bfX_{ts,k}^\ast$, $k = 1, \ldots, 6$; trained a logistic regression model using $\{\bfX_{tr}^\ast, Y_{tr}\}$; generated the predictions, $\hat{Y}_k^\ast$, using $\bfX_{ts,k}^\ast$, $k = 1, \ldots, 6$; and evaluated the respective classification accuracies across the 6 test sets.
\end{enumerate}

\subsection{Generation of the colored MNIST data}

The colored MNIST data was generated as follows: for each image in the MNIST dataset we generated a synthetic 3-channel RGB colored image, and colored it red by assigning the original grey-scale image pixels to the red channel of the synthetic images, or colored it green by assigning the original grey-scale pixels to the green channel of the synthetic image. We assigned label ``0" to digits $\{0, 1, 2, 3, 4\}$, and label ``1" to digits $\{5, 6, 7, 8, 9 \}$, and colored each digit image as either red or green. For each replication of the experiment, we colored the training set images once, while the test set images where colored in 6 different ways in order to generate the 6 distinct test sets used in the experiment. In each replication of the experiment we generated the association between the image labels and image colors by coloring a fixed proportion, $pr$, of label ``1" images as red and of label ``0" images as green, where $pr$ was set to 0.98 for the training set and the ``no shift" test set, and was set to 0.9, 0.7, 0.5, 0.3, and 0.1 for the remaining test sets (``shift 1", ``shift 2", ``shift 3", ``shift 4", and ``shift 5", respectively).

From the above description, it is clear that the data was colored according to a mediation problem (i.e., we sampled the color values, $C$, from $P(C \mid Y)$, so that the data generation process is described by the mediation model $P(\bfX \mid C, Y)P(C \mid Y)P(Y)$).

This was done, nonetheless, for computational convenience since it is just easier to color the data this way. We point out, however, that this has no impact in the validity of our results because, in anticausal prediction tasks the causality-aware approach handles confounders and mediators in exactly the same way. That is, independent of whether $C$ is a confounder or a mediator, we have that the causality-aware counterfactual features will be estimated as, $X_{j}^\ast = X_j - \hat{\beta}_{{X_j}{C}} \, C$. Hence, while in reality the approach described in this paper works to remove the influence of confounders or mediators from the features, we still present the work under the context of confounding adjustment alone because in real world situations the image data sets will be biased by confounding generated by selection biases (as described in Section 4.1) rather then biased by the influence of mediators.


\begin{thebibliography}{50}


\bibitem{rkeras2017} Allaire, J. J. and Chollet, F. (2018) keras: R Interface to 'Keras'. R package version 2.0.8.9008. https://keras.rstudio.com.

\bibitem{irm2019} Arjovsky M., Bottou L., Gulrajani I., Lopez-Paz D. (2019) Invariant risk minimization. \textit{arXiv:1907.02893v3}.

\bibitem{bareinboim2012} Bareinboim, E. and Pearl, J. (2012) Controlling selection bias in causal inference. AISTATS 2012.

\bibitem{bickel2009} Bickel, S., Bruckner, M., and Scheffer, T. (2009) Discriminative learning under covariate shift. \textit{Journal of Machine Learning Research}, \textbf{10}, 2137-2155.

%\bibitem{bot2016} Bot, B.M., et al. (2016) The mPower study, Parkinson disease mobile data collected using ResearchKit. \textit{Scientific Data} 3:160011 doi:10.1038/sdata.2016.11

%\bibitem{bottou2013} Bottou, J., et al (2013) Counterfactual reasoning and learning systems: the example of computational advertising. \textit{Journal of Machine Learning Research}, \textbf{14}, 3207-3260.

\bibitem{calders2009} Calders T., Kamiran, F., Pechenizkiy, M. (2009) Building classifiers with independency constraints. ICDM Workshop on Domain Driven Data Mining.

\bibitem{chaibubneto2019} Chaibub Neto, E., et al. (2019) Causality-based tests to detect the influence of confounders on mobile health diagnostic applications: a comparison with restricted permutations. In Machine Learning for Health (ML4H) Workshop at NeurIPS 2019 - Extended Abstract. arXiv:1911.05139.

\bibitem{achaibubneto2020a} Chaibub Neto (2020) Towards causality-aware predictions in static anticausal machine learning tasks: the linear structural causal model case. In Causal Discovery \& Causality-Inspired Machine Learning Workshop at Neural Information Processing Systems, 2020. \textit{arXiv:2001.03998}.

\bibitem{keras2015} Chollet, F. (2015) Keras, Github repository, https://github.com/fchollet/keras.

%\bibitem{chaibubneto2020} Chaibub Neto, E. (2020) Towards causality-aware predictions in static machine learning tasks: the linear structural causal model case. arXiv:2001.03998.

\bibitem{dudik2006} Dudik, M., Phillips, S. J., and Schapire, R. E. (2006) Correcting sample selection bias in maximum entropy density estimation. NeurIPS 2006.

%\bibitem{chalupka2015} Chalupka, K., Perona, P., Eberhardt, F. (2015) Visual causal feature learning. \textit{Uncertainty in Artificial Inteligence (UAI)}, 2015.

\bibitem{chawla2002} Chawla, N. V., Bowyer K. W., Hall L. O., Kegelmeyer W. P. (2002) SMOTE: synthetic minority over-sampling technique. \textit{Journal of Artificial Intelligence Research}, \textbf{16}, 321-357.

\bibitem{ghassami2017} Ghassami, A. E.,  Salehkaleybar, S., Kiyavash, N., Zhang, K. (2017) Learning causal structures using regression invariance. In \textit{NIPS 2017}.

\bibitem{gretton2009} Gretton, et al (2009)  Gretton, A., Smola, A. J., Huang, J., Schmittfull, M.,Borgwardt, K. M., and Scholkopf, B. (2009). Covariate shift by kernel mean matching.  In Quinonero-Candela,  et al., editors, \textit{Dataset  Shift  in  Machine Learning}, 131-160. The MIT Press.

\bibitem{heckman1979} Heckman, J. J. (1979) Sample selection bias as a specification error. \textit{Econometrica}, \textbf{47}, 153-161.

\bibitem{heinze2018} Heinze-Deml, C., Peters, J., Meinshausen, N. (2018) Invariant causal prediction for nonlinear models. \textit{Journal of Causal Inference}, 20170016.

\bibitem{hernan2004} Hernan, M., Hernandez-Diaz, S. and Robins, J. (2004). A structural approach to selection bias. \textit{Epidemiology}, \textbf{15}, 615-625.

\bibitem{huang2007} Huang, J., et al (2007) Correcting sample selection bias by unlabeled data. In NeurIPS 2007.

%\bibitem{johansson2016} Johansson, F. D., Shalit, U., and Sontag, D. (2016) Learning representations for counterfactual inference. \textit{International Conference on Machine Learning (ICML)}, 2017.

%\bibitem{kreif2019} Kreif, N. and DiazOrdaz, K. (2019) Machine learning in policy evaluation: new tools for causal inference. arXiv:1903.00402.

\bibitem{kamiran2012} Kamiran, F. and Calders, T. (2012)  Data preprocessing techniques for classification without discrimination. \textit{Knowledge and Information Systems}, \textbf{33}, 1-33.

%\bibitem{lopezpaz2017} Lopez-Paz, D., Nishihara, R., Chintala, S., Scholkopf, B., Bottou, L. (2017) Discovering causal signals in images. CVPR, 2017.

\bibitem{kuang2018} Kuang, K., Cui, C., Athey, S., Xiong, R., Li, B. (2018) Stable prediction across unknown environments. In \textit{SIGKDD 2018}.

\bibitem{kuang2020} Kuang, K., Xiong, R., Cui, C., Athey, S., Li, B. (2020) Stable prediction with model misspecification and agnostic distribution shift. \textit{arXiv:2001.11713}.

\bibitem{magliacane2018} Magliacane, S., van Ommen, T., Claassen, T., Bongers, S., Versteeg, P., and Mooij, J. M. (2018). Domain adaptation by using causal inference to predict invariant conditional distributions. \textit{NeurIPS 2018}.

\bibitem{pearl2009} Pearl, J. (2009) \textit{Causality: models, reasoning, and inference.} Cambridge University Press New York, NY, 2nd edition.

%\bibitem{pearl2019} Pearl, J. (2019) The seven tools of causal inference with reflections on machine learning. \textit{Communications of ACM}, \textbf{62}, 54-60.

\bibitem{peters2016} Peters, J., Buhlmann, P., Meinshausen, N. (2016) Causal inference using invariant prediction: identification and confidence intervals. \textit{Journal of the Royal Statistical Society, series B}, \textbf{78}, 947-1012.

\bibitem{quionero2009} Quinonero-Candela, J., Sugiyama, M., Schwaighofer, A., and Lawrence, N. D. (2009). \textit{Dataset shift in machine learning.} MIT Press.

\bibitem{liu2014} Liu, A. and Ziebart, B. (2014) Robust classification under sample selection bias. NeurIPS 2014.

\bibitem{shimodaira2000} Shimodaira H. (2000) Improving predictive inference under covariate shift by weighting the log-likelihood function. \textit{Journal of Statistical Planning and Inference}, \textbf{90}, 227-244.

\bibitem{rproject2019} R Core Team. (2019) R: A language and environment for statistical computing. R Foundation for Statistical Computing, Vienna, Austria. URL http://www.R-project.org/.

\bibitem{rojascarulla2018} Rojas-Carulla, M., Scholkopf, B., Turner, R., Peters, J. (2018) Invariant models for causal transfer learning. In \textit{JMLR 2018}.

%\bibitem{scholkopf2019} Scholkopf, B. (2019) Causality for machine learning. arXiv:1911.10500.

%\bibitem{schulam2017} Schulam, P., Saria, S. (2017) Reliable Decision Support Using Counterfactual Models. \textit{Neural Information Processing Systems (NIPS)}, 2017.

%\bibitem{shen2017} Shen, Z., Cui, P., Kuang, K., Li, B. (2017) On image classification: correlation vs causality. arXiv:1708.06656.

%\bibitem{sieberts2020} Sieberts, S. K., et al (2020) Crowdsourcing digital health measures to predict Parkinson's disease severity: the Parkinson's Disease Digital Biomarker DREAM Challenge. https://doi.org/10.1101/2020.01.13.904722

\bibitem{scholkopf2012} Scholkopf B, Janzing D, Peters J, et al. (2012) On causal and anticausal learning. ICML 2012, 1255-1262.

\bibitem{subbaswamy2018} Subbaswamy A., Saria, S. (2018) Counterfactual normalization: proactively addressing dataset shift and improving reliability using causal mechanisms. \textit{UAI 2018}.

\bibitem{subbaswamy2019} Subbaswamy, A., Schulam, P., Saria, S. (2019) Learning Predictive Models that Transport. \textit{AISTATS 2019}.

\bibitem{subbaswamy2020} Subbaswamy A., Saria, S. (2020) From development to deployment: dataset shift, causality, and shift-stable models in health AI. \textit{Biostatistics}, \textbf{2}, 345-352.

\bibitem{sugiyama2007} Sugiyama, M., Krauledat, M., and MAzller, K. R. (2007).   Covariate  shift  adaptation  by  importance weighted cross-validation. \textit{Journal of Machine Learning Research}, \textbf{8}, 985-1005.

\bibitem{spirtes2000} Spirtes, P., Glymour, C. and Scheines, R. (2000) \textit{Causation, Prediction and Search.} MIT Press, Cambridge, MA, 2nd edition.

%\bibitem{swaminathan2015} Swaminathan, A., and Joachims, T. (2015) Batch learning from logged bandit feedback through counterfactual risk minimization. \textit{Journal of Machine Learning Research}, \textbf{16}, 1731-1755.

%\bibitem{fashion2017} Xiao, H., Rasul, K., Vollgraf, R. (2017) Fashion-MNIST: a novel image dataset for benchmarking machine learning algorithms. arXiv:1708.07747

\end{thebibliography}
\end{document}